\documentclass[runningheads]{llncs}
\usepackage{graphicx}
\usepackage{comment}
\usepackage{caption}
\usepackage{icomma}

\usepackage[accsupp]{axessibility}

\usepackage[width=122mm,left=12mm,paperwidth=146mm,height=193mm,top=12mm,paperheight=217mm]{geometry}

\usepackage{times}
\usepackage{epsfig}
\usepackage{graphicx}
\usepackage{amsmath}
\usepackage{amssymb}
\usepackage{algorithm}
\usepackage{algpseudocode}
\usepackage{booktabs} 
\usepackage{multirow}
\usepackage{mathtools}
\usepackage{xcolor}
\usepackage[numbers]{natbib}
\usepackage{fp}

\usepackage{graphicx}
\usepackage{soul}
\usepackage{amsmath}
\usepackage{amssymb}
\usepackage{microtype}
\usepackage{wrapfig}
\usepackage{xcolor}
\usepackage{booktabs}
\usepackage{multirow}
\usepackage{xspace}

\def\figurePath{}

\def\mycfigure#1#2{%
    \begin{figure}[htb]%
    \centering\includegraphics*[width = \linewidth]{\figurePath#1}%
    \vspace{-.0cm}%
    \caption{#2}%
    \label{fig:#1}%
    \end{figure}%
}

\newcommand{\mywfigure}[3]{%
\begin{wrapfigure}{r}{#2\linewidth}%
 \vspace{-0.7cm}%
  \begin{center}%
    \includegraphics[width=\linewidth]{\figurePath#1}%
    \vspace{-.2cm}%
    \caption{#3}%
    \label{fig:#1}%
    \vspace{-1.0cm}%
  \end{center}%
\end{wrapfigure}%
}

\newcommand{\refSec}[1]{Sec.~\ref{sec:#1}}
\newcommand{\refFig}[1]{Fig.~\ref{fig:#1}}
\newcommand{\refEq}[1]{Eq.~\ref{eq:#1}}
\newcommand{\refTbl}[1]{Tbl.~\ref{tbl:#1}}
\newcommand{\refAlg}[1]{Alg.~\ref{alg:#1}}

\soulregister\ref7
\soulregister\cite7
\soulregister\refFig7
\soulregister\refAlg7
\soulregister\cite7
\soulregister\ref7
\soulregister\pageref7
\soulregister\shortcite7
\soulregister\eg0
\soulregister\ie0
\soulregister\etal0

\DeclareGraphicsExtensions{.png,.jpg,.pdf,.ai,.psd}
\newcommand{\mysection}[2]{\section{#1}\label{sec:#2}}
\newcommand{\mysubsection}[2]{\subsection{#1}\label{sec:#2}}
\newcommand{\mysubsubsection}[2]{\subsubsection{#1}\label{sec:#2}}

\newcommand{\mymath}[2]{\newcommand{#1}{\TextOrMath{$#2$\xspace}{#2}}}

\mymath{\position}{\mathbf x}
\mymath{\feature}{F}
\mymath{\layer}{l}
\mymath{\kernel}{\kappa}
\mymath{\kernelPoint}{\mathbf p}
\mymath{\expectation}{\mathbb{E}}
\mymath{\variance}{\mathbb{V}}
\mymath{\covariance}{\mathbb{C}}
\mymath{\differential}{\mathrm d}
\mymath{\offset}{\tau}
\mymath{\basis}{b}
\mymath{\estimator}{{\hat A}}

\newcommand{\method}[1]{\texttt{#1}}
\newcommand{\dataset}[1]{\textsc{#1}}

\usepackage[pagebackref,breaklinks,colorlinks]{hyperref}

\begin{document}
\pagestyle{headings}
\mainmatter
\def\ECCVSubNumber{6997}

\title{Variance-Aware Weight Initialization\\ for Point Convolutional Neural Networks}

\titlerunning{Variance-Aware Weight Initialization for Point Convolutional Neural Networks}
\author{Pedro Hermosilla\inst{1} \and
Michael Schelling\inst{1} \and
Tobias Ritschel\inst{2} \and
Timo Ropinski\inst{1}}
\authorrunning{P. Hermosilla et al.}
\institute{Ulm University, Germany \and
University College London, UK}

\maketitle

\begin{abstract}

Appropriate weight initialization has been of key importance to successfully train neural networks.
Recently, batch normalization has diminished the role of weight initialization by simply normalizing each layer based on batch statistics.
Unfortunately, batch normalization has several drawbacks when applied to small batch sizes, as they are required to cope with memory limitations when learning on point clouds.
While well-founded weight initialization strategies for regular convolutions can render batch normalization unnecessary and thus avoid its drawbacks, no such approaches have been proposed for point convolutional networks.
To fill this gap, we propose a framework to unify the multitude of continuous convolutions. This enables our main contribution, variance-aware weight initialization.
We show that this initialization can avoid batch normalization while achieving similar and, in some cases, better performance.
\end{abstract}

\section{Introduction}
\label{introduction}

Weight initialization schemes play a crucial role when training deep neural networks.
By initializing weights appropriately, diminishing or exploding layer activations can be avoided during a forward network pass. 
Accordingly, researchers have dedicated great efforts in order to optimize the weight initialization process, such that these downsides are circumvented~\cite{he2015delving}. 
As modern weight initialization schemes have been developed for structured CNNs, they are based on a few assumptions, that do not hold (as well show in \refSec{SpatialAutocorrelation}), when learning on unstructured point cloud data. 
Nevertheless, it is common practice, to overlook this discrepancy, and to apply these initialization schemes. 
These initialization schemes result in exploding or vanishing variance in layer activations, when learning on point cloud data. 
To overcome this shortcoming, usually batch normalization is applied, as it helps to rescale activations based on the current batch. 
While this approach is highly effective when learning on some moderate-resolution image data, where many training samples fit into memory and thus can be considered in one batch, it has severe downsides when learning on point clouds. 
Due to memory limitations, these data sets usually only allow for rather small batch sizes.
This makes the sample mean and variance of each individual batch usually non-representative of the entire training set. 
Therefore, in order to use batch normalization on point clouds, researchers are usually forced to reduce the point cloud resolution or to process each scene in chunks.
However, as recently shown by \citet{Nekrasov213DV} and \citet{choy20194d}, the sampling resolution and the context size play crucial roles in the final prediction of the model.
Whilst methods exist to increase the batch size virtually during training, such as accumulating the gradients over several batches or using multiple GPUs, batch normalization has the same limitations in these setups since the mean and standard deviation are computed separately in each iteration/GPU.

Within this paper, we tackle the drawbacks which result from applying weight initialization schemes, originally developed for structured data, when learning on point clouds. 
Based on our observations of layer activations in point convolutional networks, we are able to derive a variance-aware initialization scheme, which avoids the aforementioned downsides. 
To this end, we make the following contributions:

\begin{itemize}
    \item A unified mathematical framework for 3D point convolutional neural networks.
    \item We show that spatial autocorrelation increases with the depth of point convolutional networks, and show how to account for it with variance-aware weight initialization.
    \item We demonstrate how the proposed weight initialization scheme can be generalized across training data sets, and thus does not require additional preprocessing.
\end{itemize}

\noindent Since the proposed weight initialization scheme is variance-aware, we are able to omit batch normalization during the training process. 
Thus we do not only avoid the aforementioned issues which come with batch normalization, but we also are able to use larger point clouds during training and therefore arrive at en-par or sometimes even better results. 
To our knowledge, the proposed initialization scheme is the first one, which has been specifically developed for point convolutional neural networks.

\mysection{Related Work}{RelatedWork}
Training of deeper convolutional neural networks (CNNs) \cite{simonyan2014very} is a challenging optimization and hence benefits from suitable initialization.
\citet{simonyan2014very} simply used Gaussian noise with a small standard deviation.
Improving upon this, \citet{he2015delving} proposed an initialization that takes into account the effect variance of the activations.
These directly affect the numeric quality of the gradients and can be important for convergence.
Their design is specific to convolutions with certain non-linearities.
\citet{mishkin2015all} and
\citet{krahenbuhl2015data} have devised alternative inits for CNNs which proceed layer-by-layer such that that the variance of the activation affect each layer remains constant, e.g., close to one.
Alternative to good initialization, batch normalization \cite{ioffe2015batch} can serve a similar purpose: instead of changing the tunable parameters (weights), additional normalization using the statistics of each batch is employed.
These deteriorate if the sample statistics are not representative of the true statistics, which is pressing if models are complex and batches are small.
Our work aims to enable the same benefits of good initialization found for CNNs on images on unstructured convolutions \cite{atzmon2018pccnn,thomas2019KPConv,lei2019octree,hermosilla2018mccnn,wu2019pointconv,groh2018flexconv} as used for 3D point clouds.

\mysection{Formalizing Point Cloud Convolution}{convolution}
We propose a formalization to cover a large range of existing 3D point cloud convolutions.
Our contribution is to derive an initialization based on that.

Convolution of output feature $\feature^\layer$ at layer $\layer$ and position \position is defined as:
\begin{equation}
\feature^{\layer}(\position) = 
\sum_{c=0}^{C}
\int{\feature_c^{\layer-1}(\position+\offset)
\kernel_c(\offset)
\differential
\offset}
\end{equation}

\noindent where $C$ is the number of input features, $\kernel$ is the convolution kernel and \offset a spatial offset.
This formulation alone is well-known.
What is not formalized so far, however, is the many different learnable point cloud convolutions.

To represent $\kernel$, we use a $w_{c,i}$-weighted sum of the projection of the offset \offset into a set of $K$ basis functions $b_i$:
\begin{equation}
\kernel_c(\offset) = \sum_{i=0}^{K} b_i(\offset) w_{c,i},
\end{equation}

\noindent resulting in the following definition of convolution:
\begin{equation}
\feature^{\layer}(\position) = \sum_{c=0}^{C}
\int
F_c^{\layer-1}(\position+\offset)
\sum_{i=0}^{K}
\basis_i(\offset) 
w_{c,i}
\differential
\offset.
\label{eq:convolution}
\end{equation}

\noindent This definition of convolution covers many existing state-of-the-art methods, as well as the common discrete convolution used for images.
We will first show how discrete convolution can be represented by \refEq{convolution}, before applying it to continuous convolutions.

\subsection{Discrete convolution}
Here, the bases $\basis_i$ are Dirac delta functions on the positions of the kernel points $\kernelPoint_i$:
\begin{equation}
\basis_i(\offset) = \delta(\offset - \kernelPoint_i)
\end{equation}

\noindent In images, these points are placed on a regular grid in each axis.
Therefore, if pixels are laid out on a regular grid, the kernel can be reduced to a matrix indexed by pixel displacements, a summation over the neighboring pixels:
\begin{equation}
\feature^{\layer}(x) = \sum_{c=0}^{C}
\sum_{i=-\offset}^{\offset}
\feature_c^{\layer-1}(\position+i)w_{c,i}
.
\end{equation}

\subsection{Continuous convolution}
\begin{table}
    \centering
    \caption{Taxonomy of continuous convolutions.}
    \begin{tabular}{clll}
        \toprule \multicolumn1c{Method}&
         \multicolumn1c{Basis}&
         \multicolumn1c{Estim.}&
         \multicolumn1c{Init}\\
         \midrule
         \citet{atzmon2018pccnn}&
         Gauss&
         $\estimator_\mathrm{MC}$&
         \citet{he2015delving}\\
         \citet{lei2019octree}&
         Box&
         $\estimator_\mathrm{Avg}$&
         \citet{he2015delving}\\
         \citet{thomas2019KPConv}&
         Lin. Corr.&
         $\estimator_\mathrm{Sum}$&
         \citet{he2015delving}\\
         \citet{hermosilla2018mccnn}&
         MLP&
         $\estimator_\mathrm{MC}$&
         \citet{he2015delving}\\
         \citet{wu2019pointconv}&
         MLP&
         $\estimator_\mathrm{NN}$&
         \citet{he2015delving}\\
         \citet{groh2018flexconv}&
         Dot&
         $\estimator_\mathrm{Sum}$&
         \citet{he2015delving}\\
         \citet{hua2018pointwise}&
         Box&
         $\estimator_\mathrm{Avg}$&
         \citet{he2015delving}\\
         \citet{mao2019inerpo}&
         Lin. Corr.&
         $\estimator_\mathrm{Avg}$&
         \citet{he2015delving}\\
         \citet{boulch2020convpoint}&
         MLP Corr.&
         $\estimator_\mathrm{Avg}$&
         \citet{he2015delving}\\
         \bottomrule
    \end{tabular}
    \label{tbl:Zoo}
\end{table}
Several methods have been proposed to perform a convolution in the continuous domain, as it is beneficial when learning on point clouds.
As stated above, many of these methods can be expressed through the mathematical framework introduced in Section~\ref{sec:convolution}. To illustrate this, we have selected the most commonly used and highly-cited continuous convolutions to be expressed in our framework as summarized in \refTbl{Zoo}.
This ``zoo'' can be structured along two axis: the basis (\refSec{ZooBasis}) and the integral estimation (\refSec{ZooEstimation}).

\mysubsection{Zoo Axis 1: Basis}{ZooBasis}
We will here show how most published work on deep point cloud learning (those that are convolutional, which would not include the acclaimed PoinNet architecture \cite{qi2017pointnet}) can be expressed in the framework of different basis functions (\refFig{Basis}), allowing to derive a joint way of initialization.

\mycfigure{Basis}{Our framework organizes continuous convolution along the basis function (top) and  convolution integral estimation (bottom) axis.}

In \textbf{PCCNN}, \citet{atzmon2018pccnn} define the basis $b_i$ as Gaussians of the distance to a set of kernel points $\kernelPoint_i$:
\begin{equation}
\basis_i(\offset) 
=
\exp
\left(
-\frac
{\|\kernelPoint_i - \offset\|^2}
{s}
\right),
\end{equation}
where $s$ is a bandwidth parameter.

For \textbf{PointWiseCNN}, \citet{hua2018pointwise} also used a set of kernel points as PCCNN. However, the authors used a box function as point correlation:
\begin{equation}
\basis_i(\offset) =
    \begin{cases}
      1 & \underset{j}{\arg\!\min}(\|\kernelPoint_j - \offset\|) = i\\
      0 & \underset{j}{\arg\!\min}(\|\kernelPoint_j - \offset\|) \neq i
    \end{cases}
\end{equation}

This approach was also adopted by \citet{lei2019octree} in \textbf{SPHConv}, but in spherical coordinates.

In \textbf{KPConv}, \citet{thomas2019KPConv} also used a set of kernel points as PCCNN. However, the authors used linear correlation instead of a Gaussian:
\begin{equation}
\basis_i(\offset) = 
\max
\left(
1 - \frac{\| \kernelPoint_i - \offset\|}{s}, 0
\right).
\end{equation}

\noindent Here, kernel points are arranged as the vertices of platonic solids.

Linear correlation was also used by \citet{he2015delving} in their convolution operation \textbf{InterpCNN} with kernel points arranged in a grid.

In \textbf{ConvPoint}, \citet{boulch2020convpoint} also used kernel points. 
However, the correlation function was learned by an MLP instead.

In the \textbf{MCConv} work of \citet{hermosilla2018mccnn}, the basis $b_i$ is defined as the output of an MLP $\alpha(\offset)$ with a structure too complex to be written as an equation here.
According to our taxonomie's axes, \textbf{PointConv} by \citet{wu2019pointconv} is the same as MCConv and only differs in the implementation of the MLP and along the integration design axis.

The \textbf{FlexConv} approach by \citet{groh2018flexconv} uses a single basis $\mathbf v_i$ for each input feature and it is defined as the affine projection $\cdot_1$ of the point to the learned vector:
\begin{equation}
\basis_i(\offset) = \mathbf v_i^T \cdot_1 \offset
.
\end{equation}

\noindent This basis can be interpreted as learned unit vectors scaled by the convolution weight.

\mysubsection{Zoo Axis 2: Integral Estimation}{ZooEstimation}
Orthogonal to the choice of basis just discussed, different methods also use different ways of estimating the inner convolution integral in \refEq{convolution}.
To see those differences, consider writing the integral as
\begin{align}
\label{eq:ConvolutionAsIntegral}
A(\position)
=&
\int a(\position,\offset) \differential \offset\text{\quad where}\\
a(\position,\offset) 
=& 
F_c^{\layer-1}(x+\offset)
\sum_{i=0}^{K} b_i(\offset)
\mathbf w_{c,i}.\nonumber
\end{align}

In point cloud processing, several  ways have been proposed to estimate this integral, based on summing, averaging, MC estimation and MC with learned density:

\begin{alignat}{2}
\label{eq:int_sum}
\estimator_\mathrm{Sum}(\position) =& 
\sum_{y\in\mathcal{N}(\position)}
a(y)
&&\text{\ \cite{groh2018flexconv, thomas2019KPConv}},\\
\label{eq:int_n}
\estimator_\mathrm{Avg}(\position) =& 
\sum_{y\in\mathcal{N}(\position)}
a(y)/|\mathcal{N}(\position)|
&&\text{\ \cite{hua2018pointwise, lei2019octree},}\\
\label{eq:int_n_pdf}
\estimator_\mathrm{MC}(\position) =& 
\sum_{y\in\mathcal{N}(\position)}
a(y)/
(p(y)|\mathcal{N}(\position)|)
&&\text{\ \cite{atzmon2018pccnn, hermosilla2018mccnn},}\\
\label{eq:int_n_nn}
\estimator_\mathrm{NN}(\position) =& 
\sum_{y\in\mathcal{N}(\position)}
a(y)/
\pi(p(y))
&&\text{\ \cite{wu2019pointconv}.}
\end{alignat}

The most similar approach to the discrete convolution case is to use a sum (\refEq{int_sum}) over the neighboring samples~\cite{groh2018flexconv, thomas2019KPConv}.
Although easy to implement, it is sensitive to neighborhoods with a variable number of samples, as well as to non-uniformly sampled point clouds (second row in \refFig{Basis}).
To consider this shortcoming, other approaches normalize over the average (\refEq{int_n}) of the neighboring contributions~\cite{hua2018pointwise, lei2019octree, mao2019inerpo, boulch2020convpoint}.
This is robust to neighborhoods with a different number of samples.
However, they are not robust under non-uniformly sampled point clouds.
To be able to learn robustly on non-uniformly sampled point clouds, other methods have used a weighted sum \refEq{int_n_pdf}, where weights depend on the density of the points, following the principles of Monte Carlo integration~\cite{atzmon2018pccnn, hermosilla2018mccnn}.
\citet{wu2019pointconv} additionally propose an MPL $\pi$ to map the estimated density to a corrected density as per \refEq{int_n_nn}.
These methods are robust under different numbers of neighboring samples, as well as non-uniformly sampled point clouds.

\section{Weight Initialization}
\label{weightinit}

We will now derive weights $w$  that are \emph{optimal in terms of feature variance} for the general form of convolution we describe  in  \refEq{Convolution}, for any form of convolution integral estimation and any basis.

Weights are optimal, if the variance of the features does not increase/decrease for increasing layer depth \cite{he2015delving}.
This is best understood from plots where the horizontal axis is network depth and the vertical axis is variance (\refFig{Motivation}).
A method, such as uniform initialization will have increasing/decreasing variance for increasing depth.
Previous work \cite{he2015delving} has enabled keeping the variance constant as shown by the pink curve.
All the continuous methods use similar initializations where variance decreases (green curve).
Our contribution is the blue curve: variance remains constant for the continuous case.
\mywfigure{Motivation}{0.5}{Good initialization (red and blue) prevents decreasing variance (vertical axis) with increasing layer depth (horizontal axis).}

Based on the convolution framework introduced in Section~\ref{sec:convolution}, we will in this section, first describe the weight initialization commonly used for discrete convolutions, before detailing the currently used weight initialization for point convolutional neural networks. Based on the shortcomings of these initialization schemes, we will further introduce our new initialization scheme, which we have developed for point convolutional neural networks.

\subsection{Discrete convolutions}
The weight parameters of a discrete convolution are usually initialized independently of the input data. 
The underlying idea is to initialize the weights in such a way, that the variance of the output features is the same as the variance of the input. 
The weights are therefore initialized using a normal distribution with a carefully selected variance. 
This variance is computed, by relying on several assumptions.
First, that the weights are independent of each other, and second, that the features of each pixel are independent of each other. Following these assumptions, \citet{he2015delving} derived the appropriate variance of the weights for convolutional neural networks with ReLUs:

\begin{equation}
\variance[w] = \frac{2}{N C}
,
\end{equation}

\noindent where $N$ is the number of pixels in the kernel, and $C$ is the number of input features.

\subsection{Continuous convolutions}
In this section, we discuss the implications arising from applying classical weight initialization approaches in the context of point convolutional neural networks, and propose our approach, specifically tailored to point convolutional neural networks.

\subsubsection{Common practices}
A na\"ive approach would be to use the same initialization scheme for continuous convolution as the one used for discrete convolutions.
However, the number of neighboring points highly depends on the data to process, e.g., convolution on samples on a plane will have fewer neighbors than convolutions on molecular data.
Therefore, a common approach is to rely on the standard weight initialization schemes provided by the software packages, which in the best case results in the following variance:

\begin{equation}
    \variance[w] = \frac{2}{B C}
    \label{eq:std},
\end{equation}

\noindent where $B$ is the number of basis functions and $C$ is the number of features. 
Other methods, such as presented by \citet{hermosilla2018mccnn} and \citet{thomas2019KPConv}, in their implementation, simply divide by $C$ which produces an even more biased variance and, therefore, worse results.
We will consider \refEq{std} as the standard initialization.

\mysubsubsection{Spatial autocorrelation}{SpatialAutocorrelation}
Even though approximating the number of neighbors by the number of basis functions is a crude approximation, this initialization scheme is designed using some assumptions that do not hold, when considering the continuous case.
The derivations for the discrete case assumed that features from neighboring pixels are independent of each other.
In the continuous case, however, they are correlated.

\mycfigure{Correlation}{
\textbf{a}: Minimal setting for continuous convolution: Starting from the top 1D point cloud we convolve with a 1D  3-tap box kernel 20 times.
\textbf{b}: Spatial autocorrelation of features in each layer as a function of the distance between points this setting.
The top row shows that there is no spatial autocorrelation for the discrete case. However, for continuous convolutions, we can see high spatial autocorrelation for close points and an increase with the depth of the network.}

\refFig{Correlation} shows the empirical correlogram of the features of different layers of a point convolutional neural network.
To obtain this data, we used the simplest 1D point cloud and the simplest continuous convolution conceivable (\refFig{Correlation},a):
As a point cloud we sample positions from the uniform random distribution on $(0,1)$, and the initial features from a normal distribution with variance $0.1$.
As a basis we use three boxes separated by $r=.05$ and \refEq{int_sum} to estimate the integral.
 \refFig{Correlation},b depicts the correlation as a function of the distance between points for different layer depth.
We can see that no clear pattern for the spatial autocorrelation in the initial features (Input) emerges.
However, after the initial convolution, the correlation increases for close points, and this correlation slightly increases and widens with the depth of the network.
This is empirical evidence, how the assumption of independence of features between neighbors does not hold in the continuous case, and thus an initialization scheme rooted on this might be suboptimal.

\subsubsection{Variance-aware weight initialization}
To obtain a more suitable weight initialization scheme for point convolutional neural networks, we start our derivation with the definition of the variance of layer $l$:
\begin{equation}
\variance[\feature^{\layer}(\position)] = 
\expectation[\feature^{\layer}(\position)^2] - 
\expectation[\feature^{\layer}(\position)]^2
\label{eq:layer_variance}
\end{equation}
\noindent We will perform the derivations for a single input channel and use the assumption that each feature channel is independent of each other to scale the resulting variance by the number of channels, $C$.

Starting with \refEq{layer_variance}, we compute the expectation of the output features of layer $l$. Therefore, we assume that each weight $w_i$ is independent of each other, as well as from the basis functions and features, and that they are further initialized from a normal distribution centered at $0$. Accordingly, we can reformulate the expectation of the output features of layer $l$ as follows:
\begin{align}
\expectation[\feature^{\layer}(\position)] 
&= 
\expectation
\left[
\int
\feature^{\layer-1}(x+\offset) \sum_{i=0}^{K} b_i(\offset) 
w_{c,i} 
\differential
\offset
\right]
\nonumber\\
&=
\int
\sum_{i=0}^{K} \expectation
\left[
\feature^{\layer-1}(\position+\offset) 
\basis_i(\offset)
\right]
\expectation\left[w_{c,i}
\right] 
\differential
\offset \nonumber\\
&=
\int
\sum_{i=0}^{K} \expectation
\left[\feature^{\layer-1}(x+\offset) \basis_i(\offset)
\right]
0
\differential
\offset
\nonumber\\
&= 0\nonumber
\end{align}

\noindent As the expectation equals $0$, the variance defined in \refEq{layer_variance} has to be equal to the expectation of the squared features:
\begin{equation}
    \variance[\feature^{\layer}(\position)] = 
    \expectation[\feature^{\layer}(\position)^2] \nonumber
\end{equation}

\noindent If we expand $\expectation[\feature^{\layer}(\position)^2]$ with our definition of convolution we obtain that it is equal to
\begin{equation}
\label{eq:Convolution}
    \expectation
    \left[ 
    \int
    G_{1} 
    \sum_{i=0}^{K} H_{i,1} w_{k,i} 
    \differential
    \offset_1
    \int
    G_{2} 
    \sum_{j=0}^{K} H_{j2} w_{l,j} 
    \differential
    \offset_2
    \right],
    \nonumber
\end{equation}

\noindent where $G_{1/2} = \feature^{\layer-1}(x+\offset_{1/2})$ and $H_{i1/2} = b_i(\offset_{1/2})$. Since $w$ is independent of the features and basis functions, we can re-arrange the terms to obtain
\begin{equation}
    \int 
    \int
    \sum_{i=0}^{K}
    \sum_{j=0}^{K}
    \expectation
    \left[
    G_{1} G_{2} H_{i1} H_{j2}
    \right]
    \expectation
    \left[
    w_{ki} w_{lj} 
    \right]
    \differential \offset_1 
    \differential \offset_2
    .
    \nonumber
\end{equation}

\noindent This equation can be simplified using the assumption that all weights also are independent of each other, and that the weights are drawn from a distribution centered at $0$:
\begin{align}
   \sum_{i=0}^{K}
   \sum_{j=0}^{K}
   \expectation
   \left[w_i w_j \right] 
   &= 
   \sum_{i=0}^{K} 
   \sum_{j=0}^{K}
   \covariance[w_i, w_j] + 
   \expectation[w_i] 
   \expectation[w_j] 
   \nonumber = \sum_{i=0}^{K} 
    \variance[w] 
    + 0
    \nonumber
    .
\end{align}
\noindent We can reformulate $\expectation[\feature^{\layer}(\position)^2]$ as:
\begin{equation}
   \int \int \sum_{i=0}^{K}
   \expectation
   \left[
   G_{1} G_{2} H_{i,1} H_{i,2}
   \right]
   \variance
   \left[w\right]
   \differential \offset_1
   \differential \offset_2
   \nonumber
   .
\end{equation}

\noindent Finally, we can arrange the terms to isolate the variance of the weights $w$ on one side of the equality, and thus obtain a closed form to determine the optimal variance for our weights. When additionally including the constant $C$, to account for the number of input features, we obtain:
\begin{align}
   \variance[w] =& \frac{\variance[\feature^{\layer}(\position)]}{C z_l}
   \text{\qquad where }
   \label{eq:var_w}
   \\
   z_l =& \expectation
   \left[
   \sum_{i=0}^{K}
   \int G_{k2}H_{i2}
   \int G_{k1}H_{i1}
   \differential \offset_1
   \differential \offset_2
   \right]
   .
   \label{eq:denom}
\end{align}

\noindent This accounts for the spatial autocorrelation of the input features ($G_{k1} G_{k2}$) as well as the basis functions ($H_{i1} H_{i2}$).

\mysubsection{Variance computation}{Variance Computation}
In order to apply this initialization, we have to compute \refEq{var_w}.
First, we specify the desired variance in the output of each layer, $\variance[\feature^{\layer}(\position)]$, which can be chosen to be a constant such as $1$.
This leaves us with computing \refEq{denom}, a statement for layer $l$ given the features in the previous layer $l-1$.
Hence, the computation proceeds by-layers, starting at the first layer, adjusting variance and proceeding to the next layer as seen in the outer loop in \refAlg{data_driven_alg}.

\begin{wrapfigure}{l}{0.47\textwidth}
    \hrule height 1pt%
    \vspace{4pt}
    \captionof{algorithm}{Weight initialization.}%
    \label{alg:data_driven_alg}%
    \vspace{-6pt}
    \hrule%
    \vspace{4pt}
    \begin{algorithmic}[1]
        \For{$l \in \mathrm{Layers}$}
            \State $z_l \leftarrow 0$
            \For{$1$ to $N$}
            \Comment \refEq{denom}
                \State $\mathcal P \leftarrow \mathtt{samplePointCloud()}$
                \State $\bar z_l \leftarrow\mathtt{ estimate}(\mathcal P,l)$
                \State $z_l \leftarrow z_l + \bar z_l$
            \EndFor
            \State $z_l \leftarrow z_l/N$
        \EndFor
    \end{algorithmic}
    \vspace{4pt}
    \hrule
\end{wrapfigure}
The outer expectation in \refEq{denom} is estimated by sampling $N$ random point clouds $\mathcal P$.
The inner double-integral for each point cloud is a double convolution with the same structure as the convolution \refEq{ConvolutionAsIntegral}.
Hence, it can be estimated using one of the techniques defined by \refEq{int_sum}, \refEq{int_n} or \refEq{int_n_pdf}.
Depending on which technique is used, this might or might not be an unbiased estimator of the true convolution integral, but at any rate, for the initialization to work, the same estimation has to be used that will later be used for actual learning.
The function $\mathtt{estimate}(\mathcal P,l)$ executes the estimation with the weights already initialized up to level $l-1$ on the point cloud $\mathcal P$.

As we will show later, this algorithm can be used to estimate $z_l$ from the training data for a specific architecture or can be pre-computed from a set of representative point clouds.
In the latter case, the $z_l$ is estimated for a set of consecutive layers where each will capture the spatial autocorrelation introduced by previous layers.
These $z_l$ values can be later queried for another network architecture based on the depth of each layer, scaling them by the number of input features, and use the result to initialize the layer.


\section{Experiments}
\label{experiments}

\subsection{Operators}
In order to evaluate the proposed initialization scheme, we have selected widely used point convolution approaches, that can be expressed in our framework introduced in Section~\ref{sec:convolution}. During the selection process, we have ensured, that we cover a variety of different basis functions as well as integral approximations and have selected
\method{PCCNN}~\cite{atzmon2018pccnn}, \method{KPConv}~\cite{thomas2019KPConv}, \method{SPHConv}~\cite{lei2019octree}, \method{MCConv}~\cite{hermosilla2018mccnn}, and \method{PointConv}~\cite{wu2019pointconv}.

\subsection{Variance evaluation}
\begin{figure}[!t]%
\centering\includegraphics*[width = \linewidth]{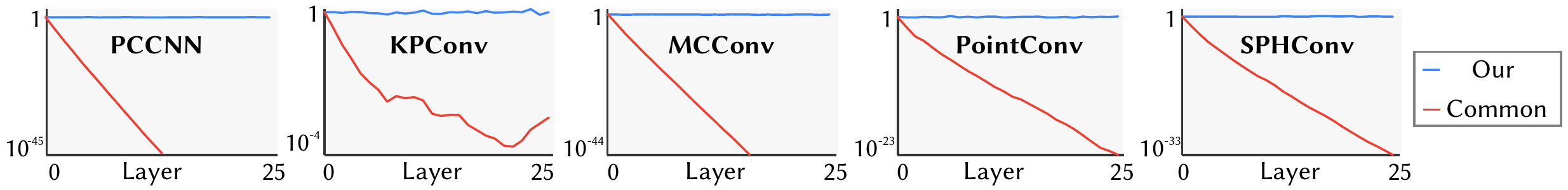}%
\caption{Comparison of the variance of each layer when applying standard weight initialization (\emph{red}), and when applying our weight initialization (\emph{blue}). The plots show the results for different point convolution operators.
Note the vertical axis to be in log-scale.}
\label{fig:variance}%
\end{figure}

To empirically validate our derivations, we compare the variance of each layer in a $25$-layer network initialized with the standard initialization (\refEq{std}) and initialized with ours (\refEq{var_w}). \refFig{variance} shows the results obtained from this experiment. As we can see, while the variance of each layer exponentially approaches zero for the standard initialization, ours maintains a constant variance over all layers for all tested convolution operators.

\subsection{Classification}

We validate our algorithm on the task of shape classification on the \dataset{ScanObjectNN} data set~\cite{uyscanobjectnniccv19}. 
Since all shapes contain similar numbers of points, this is an ideal setup to observe the effects of our initialization scheme under varying batch sizes.

\noindent \textbf{Data set: }
The data set is composed of $2,902$ real objects from $15$ different categories, obtained from real 3D scans of the \dataset{ScanNet} data set~\cite{dai2017ScanNet}.
For this experiment, we use the data set in which only points from the object are used as input to the network, i.e., the background points have been removed.
We use the official training/test split which uses $80$\,$\%$ objects for training and $20$\,$\%$ for testing.
We sample $1,024$ random points from each object from the $2,048$ points provided in the data set.
As initial point features, we use a single float with the value of $1.0$, while we use random anisotropic scaling on the point coordinates (random scaling in each axis independently between $0.9$ and $1.1$) and random rotation along the vertical axis for data augmentation.
Performance is measured as overall accuracy accumulating the predicted probabilities for each model with different data augmentations.

\mywfigure{Architecture}{0.5}{Network architectures used for a semantic segmentation task on \dataset{ScanNet} (top) and for a classification task on \dataset{ScanObjectNN} (bottom).}

\noindent \textbf{Network architecture: }We used an encoder network with three resolution levels, and two convolution layers per level (see \refFig{Architecture}). 
In order to compute the point cloud levels, we used Poisson disk sampling with radii [$0.1, 0.2, 0.4$], and three times this radius as the convolution's receptive field. 
This results in a total of six convolution layers. 
We also used an increasing number of features per level, [$128, 256, 512$], and a global average pooling layer on the resulting features to create the global shape descriptor. 
This descriptor was then processed by a single layer MLP with $512$ hidden units, which generates the resulting probabilities. 
In order to have the same number of parameters for all different convolution operations, we used $16$ basis functions for all of them, which enables analyzing the effect of our initialization scheme, but avoids overfitting issues.

\noindent \textbf{Training: }We trained the models using SGD with momentum for $650$ epochs, batch size between $2$ and $16$, and an initial learning rate of $0.005$. To enable convergence for all methods, we scaled the learning rate by $0.1$ after $500$ epochs. In order to prevent overfitting, we used a dropout value of $0.2$ before each convolution and $0.5$ on the final MLP, and we used weight decay loss scaled by $0.0001$.

\begin{table}
\setlength{\tabcolsep}{1.75pt}
\begin{center}
\captionof{table}{Comparison of ours and standard initialization, Group Normalization, and batch normalization on the \dataset{ScanObjectNN} data set for different batch sizes.}
\label{tbl:scanobjnn}
\begin{tabular}{crrrrrrrrrrrrrrrrr}
    \toprule
    Batch size $\rightarrow$ & \multicolumn{5}{c}{2}&&\multicolumn{3}{c}{4}&&\multicolumn{3}{c}{8}&&\multicolumn{3}{c}{16}\\
    \cmidrule(lr){2-6}\cmidrule(lr){8-10}\cmidrule(lr){12-14}\cmidrule(lr){16-18}
    & Ours & He & Xav. & GN & BN && Ours & GN & BN && Ours & GN & BN && Ours & GN & BN\\
    \midrule
    PCCNN~\cite{atzmon2018pccnn} & 85.7 & 13.5 & 13.5 & \textbf{86.1} & 23.3 & & 85.5 & 86.7 & \textbf{87.3} && 85.2 & 86.7 & \textbf{87.3} && 83.5 &  86.6 & \textbf{87.9} \\ 
    KPConv~\cite{thomas2019KPConv} & \textbf{84.8} & 78.9 & 13.5 & 83.8 & 17.7 && \textbf{84.5} & 83.8 & 83.4 && \textbf{85.1} & 84.6 & 84.8 && 84.1  & 84.1 & \textbf{85.3} \\ 
    MCConv~\cite{hermosilla2018mccnn} & \textbf{86.3} & 13.5 & 13.5 & 85.5 & 21.3 && \textbf{85.9} & 85.7 & 83.5 && 85.7 & 85.6 & \textbf{85.9} && 84.4 & 85.0 & \textbf{85.6} \\ 
    PointConv~\cite{wu2019pointconv} & \textbf{85.7} & 85.1 & 85.3 & 85.0 & 20.1 && \textbf{85.2} & 84.8 & 83.4 && 85.4 & 84.6 & \textbf{85.9} && \textbf{85.4} & 84.9 & \textbf{85.4} \\ 
    SPHConv~\cite{lei2019octree} & \textbf{82.2} & 13.5 & 13.5 & 79.8 & 26.2 && \textbf{81.9} & 80.8 & 81.6 && 81.0 & 80.3 & \textbf{82.8} && 81.0 & 80.8 & \textbf{83.4} \\ 
    \bottomrule
\end{tabular}
\end{center}
\end{table}

\noindent \textbf{Results: } The resulting accuracy for different methods and different batch sizes is shown in Table~\ref{tbl:scanobjnn}. 
We see that our initialization allows to eliminate batch normalization from the network without a significant decrease in performance for most convolution operators.
Moreover, we can see that for low batch sizes, our initialization obtains better performance than batch normalization, whose performance reduces with the batch size.
Lastly, we can see that our initialization scheme outperforms batch normalization for some convolution operators (\method{MCConv}), where a small batch size acts as regularization during training.
When compared to standard initialization schemes, such as He~\cite{he2015delving} or Xavier~\cite{pmlrv9glorot10a}, our method always obtains better performance while these methods are not able to converge for most of the convolution operators.
Moreover, we compare our initialization scheme with Group Normalization~\cite{Wu_2018_ECCV}, a common normalization technique used for small batch sizes.
Tbl.~\ref{tbl:scanobjnn} shows that, although Group Normalization enables network training with small batch sizes, with most convolution operators and batch sizes, our initialization scheme obtains better results.

\subsection{Semantic Segmentation}

We also evaluated our method on the task of semantic segmentation of the \dataset{ScanNet} data set~\cite{dai2017ScanNet}.
In this task, each scene is too big to be processed as a whole and it has to be processed in blocks, allowing us to validate our initialization for different scene sizes.

\noindent \textbf{Data set:} The \dataset{ScanNet} data set~\cite{dai2017ScanNet} is composed of real 3D scans from $1,513$ different rooms, where the network has to predict the class of the object to which each point belongs to. 
We use the official splits, corresponding to $1,045$ rooms for training, $156$ rooms for validation, and $312$ rooms for testing.
Since the ground truth annotation for the test set is not publicly available, we evaluated the methods on the validation set.
We sample each scan with a sphere of radius $r_1$ around a random point in the scene and using another sphere of bigger radius, $r_2$, to select the context, i.e., the points of the bigger sphere are input to the network but we only perform predictions for the points inside the small sphere.
We use two sets of radii in our experiments.
First, we use $r_1 = 2$\,m and $r_2 = 4$\,m, resulting in point clouds of around $120$\,k points for the bigger sphere and around $45$\,k for the smaller one.
In this setup, we fill the available memory in our system and use only a point cloud from a single room in each training step.
Then, we also use $r_1 = 1$\,m and $r_2 = 2$\,m, resulting in smaller point clouds of around $12$\,k and $45$\,k points.
These radii provide a smaller context for the prediction but allow us to process four rooms in each batch.
We use random rotation along the up vector and anisotropic scaling to augment the data during training.
Performance is measured as Intersection Over Union (IoU), by accumulating the predicted probabilities of each point sampled from different spheres and with different random rotations and scalings.

\noindent \textbf{Network architecture: }We used a U-Net architecture~\cite{ronneberger2015u} with four different levels and with two convolution layers in each level (see \refFig{Architecture}). 
The different levels are computed using Poisson Disk sampling with different radii, $[0.03, 0.06, 0.12, 0.24]$, and the receptive field of the convolution layers uses three times the Poisson radius of the level. 
The number of features increases per level in the encoder and decreases per level in the decoder, resulting in [$64, 128, 256, 512, 256, 128, 64$].
The up-sampling operations in the decoder are also continuous convolutions, which results in $19$ convolution layers in total.
As before, we used $16$ basis functions for all tests and all methods. 

\noindent \textbf{Training: }We trained the models using SGD with momentum for $500$ epochs, and, for each epoch, we sample $3000$ point clouds.
We used an initial learning rate of $0.005$, which was scaled by $0.25$ after $300$ and again by $0.25$ after $400$, allowing all methods to converge.
In order to prevent overfitting, we used a dropout value of $0.2$ before each convolution and weight decay loss scaled by $0.0001$.

\begin{wraptable}{r}{0.65\textwidth}
\setlength{\tabcolsep}{3pt}
\centering
\caption{Comparison of our initialization on different convolution operations, for Semantic Segmentation on \dataset{ScanNet}, on different batch sizes.}
\label{tbl:segment}
\begin{tabular}{crrrrrr}
    \toprule
    \multicolumn1r{Batch size $\rightarrow$}&
    \multicolumn{4}{c}{1}&
    \multicolumn{2}{c}{4}\\
    \cmidrule(lr){2-5}\cmidrule(lr){6-7}
    \multicolumn1r{Batch norm $\rightarrow$}&
    \multicolumn{1}{c}{Yes}&
    \multicolumn{3}{c}{No}& 
    \multicolumn{1}{c}{Yes}&
    \multicolumn{1}{c}{No}\\
    \cmidrule(lr){2-2}
    \cmidrule(lr){3-5}
    \cmidrule(lr){6-6}
    \cmidrule(lr){7-7}
    &  &  & \multicolumn{2}{c}{Ours} &  & \multicolumn{1}{c}{Ours}\\
    \cmidrule(lr){4-5}\cmidrule(lr){7-7}
    & Std & Std & Dir. & Tran. & Std & Dir.\\
    \cmidrule{1-7}
    PCCNN~\cite{atzmon2018pccnn} & 
        32.4 & 13.3 & \textbf{65.1} & 63.6 & \textbf{65.7} & 62.6\\ 
    KPConv~\cite{thomas2019KPConv} & 
        33.9 & 52.5 & \textbf{66.1} & 64.9 & 66.0 & \textbf{67.7}\\
    MCConv~\cite{hermosilla2018mccnn} & 
        40.3 & 58.8 & 67.4 & \textbf{67.9} & \textbf{66.1} & 65.1\\ 
    PointConv~\cite{wu2019pointconv} & 
        41.4 & 1.6 & \textbf{62.3} & 61.7 & \textbf{64.2} & 62.5\\
    SPHConv~\cite{lei2019octree} & 
        28.1 & 46.4 & \textbf{60.7} & 60.6 &59.5 & \textbf{60.1}\\
    \bottomrule
\end{tabular}%
\end{wraptable}%
\noindent \textbf{Results: } When we train the models with a single scene per batch, Table~\ref{tbl:segment} shows that batch normalization is not able to successfully learn the task.
Our initialization scheme, on the other hand, enables us to eliminate batch normalization obtaining competitive IoUs. 
Moreover, we can see that for some methods, such as \method{KPConv}, \method{MCConv}, and \method{SPHConv}, using standard initialization without batch norm can result in an improvement on performance wrt. batch norm.
However, these results do not match the IoU obtained with our initialization scheme.
This is true for both variants of ours, the ``direct'' and the ``transfer'' one, as shown in the table.
The direct one will perform the initialization based on access to the data to choose weights optimally.
For the ``transfer'' one, we computed \refEq{denom} on synthetic shapes from the \dataset{ModelNet40} data set~\cite{wu2015modelnet} on a network with $25$ consecutive layers without any pooling operation or skip connections.
Then, we used this initialization on the network for the semantic segmentation task.
The only additional step is to select the appropriate $w_l$ value based on the layer depth and scale this value by the number of input channels $C$ of the layer.
We see that this transfer of weight initialization can work, as performance remains similar while retaining all the benefits and avoiding the need to do anything when deploying our initialization for a specific architecture.
Table~\ref{tbl:segment} also shows the results obtained when the networks are trained with four small point clouds per batch.
We can see that in this setup, batch normalization achieves high IoU.
However, for some methods, \method{MCConv} and \method{SPHConv}, using larger point clouds with our initialization significantly increases the obtained IoU.


\section{Limitations}
\label{limitations}
Our method is not exempt from limitations, the main one being, that, contrary to most initialization schemes used for CNN in images, it is data-dependent and requires processing the data first.
This pre-processing overhead is negligible to the overall training time: for the \dataset{ScanNet} task results in $47.5$ additional seconds from a total of $3$ days of training, and for the \dataset{ScanObjectNN} task results in $6.1$ additional seconds from a total of $4$ hours of training.
However, we also propose a method to transfer weight initialization's between tasks that does not require pre-processing the data, i.e., our ``transfer'' setup.
However, a large difference between the shapes used to transfer the weight initialization and different statistics between the input features of the data sets could lead to low performance for some operators.

\section{Conclusions}
\label{conclusions}
In this paper, we have introduced a novel, variance-aware weight initialization scheme, developed for point convolutional neural networks. By exploiting spatial autocorrelation within the layers of a point convolutional neural network, we were able to derive the weight variance used for initialization. 
In contrast to standard weight initialization schemes, which have been developed for structured data, our proposed weight initialization scheme allows omitting batch normalization, which leads to several issues in the learning process. 
We have shown, that when using our weight initialization scheme, we are able to neglect batch normalization, and still obtain the same or sometimes even better learning performance. 
Moreover, we have shown that our weight initialization allows training with small batch sizes and, therefore, larger point clouds, the main limitation of using batch normalization on point clouds.
We believe that the proposed weight initialization scheme is the first one developed for point convolutional neural networks, and we hope that it will establish itself as the standard method in this subfield.

In the future, we see several endeavors for future work.
Same as \citet{he2015delving}, we assume features to be independent, which would not be true for positions and normals (one is approximately the derivative of the other) or position or orientation and color (due to shading).
Further, we would like to investigate the impact of our weight initialization scheme in other domains, such as graphs and problems such as protein learning.
\\
\\
\noindent \textbf{Acknowledgements.}
{
This work was partially funded by the Deutsche Forschungsgemeinschaft (DFG) under grant 391088465 (ProLint) and by the Federal Ministry of Health (BMG) under grant ZMVI1-2520DAT200 (AktiSmart-KI).
}

\clearpage

{\small
\bibliographystyle{splncs04nat}
\bibliography{main}
}

\clearpage

\appendix

\section{Additional experiments}
\label{imp}

\subsection{ModelNet40}

We validate our algorithm on the task of shape classification on the \dataset{ModelNet40} data set~\cite{wu2015modelnet}, composed of only synthetic 3D models. 

\noindent \textbf{Data set: }
This synthetic data set is composed of $12$\,k shapes from $40$ different human-made objects, whereby the shapes are aligned into a canonical orientation.
We use the official training/test split which uses $9.8$\,k shapes for training and $2.4$\,k shape for testing.
We sample points using Poisson disk sampling with a radius of $0.05$ from $2$\,k initial random points on the surface of the objects, which results in roughly $1$\,k points per model.
As features, we use the surface normals, while we use random anisotropic scaling (random scaling in each axis independently between $0.9$ and $1.1$) for data augmentation.
Performance is measured as overall accuracy accumulating the predicted probabilities for each model with different anisotropic scalings.

\noindent \textbf{Network architecture: }We used the same architecture as the one used on the \dataset{ScanObjectNN} data set, increasing the number of features by a factor of $2$ and decreasing the sampling radius and convolution receptive fields by a factor of $2$.

\noindent \textbf{Training: }We trained the models using SGD with momentum for $500$ epochs, batch size equal to $16$, and an initial learning rate of $0.005$.
We scaled the learning rate by $0.1$ after $350$ and again by $0.1$ after $450$, allowing all methods to converge.
In order to prevent overfitting, we used a dropout value of $0.2$ before each convolution and $0.5$ on the final MLP, and we used weight decay loss scaled by $0.0001$.

\noindent \textbf{Results: } The resulting accuracy is shown in Table~\ref{tbl:class}. 
Our initialization scheme enables us to eliminate batch normalization while still achieving accuracy values comparable to the standard initialization with batch normalization. In contrast, when using standard initialization without batch normalization, three tested approaches do not learn.
This is true for both variants of ours, the ``direct'' and the ``transfer'' one, as shown in the last two columns.
For the ``transfer'' one, we computed the initialization on the \dataset{ScanNet} data set~\cite{dai2017ScanNet} for every method on a network with six consecutive layers without any pooling operation.
Then, we used this init on the network for the classification task.
We see that the performance remains the same as the ``direct'' setup.

Lastly, we compare our initialization with batch norm when the batch size is reduced. 
In this experiment, we trained our classification network with and without batch normalization for batch sizes $2$, and $4$. 
The results, shown in Table \ref{tbl:class_small_batch}, indicate that for a small batch size of $2$ our method is able to train the model while methods using batch normalization are not able to converge.

\begin{table}
\begin{minipage}[t]{0.475\textwidth}

\centering
\setlength{\tabcolsep}{4pt}
\caption{Comparison of accuracy obtained with our initialization on different convolution operations, for \dataset{ModelNet40} Classification.}
\label{tbl:class}
\begin{tabular}{crrrr}
    \toprule
    \multicolumn1r{Batch norm $\rightarrow$}& Yes & \multicolumn{3}{c}{No}\\
    \cmidrule{3-5}
    &  &  & \multicolumn2c{Ours} \\
    \cmidrule{4-5}
    &
    \multicolumn1c{Std}&
    \multicolumn1c{Std}&
    \multicolumn1c{Dir.}& \multicolumn1c{Tran.}\\
    \cmidrule{1-5}
    PCCNN~\cite{atzmon2018pccnn} & 
        \textbf{91.0} & 4.1 & 90.0 & 89.7\\ 
    KPConv~\cite{thomas2019KPConv} & 
        91.1 & 90.8 & 91.0 & \textbf{91.2}\\
    MCConv~\cite{hermosilla2018mccnn} & 
        \textbf{91.0} & 4.1 & 90.4 & 90.6\\ 
    PointConv~\cite{wu2019pointconv} & 
        90.7 & \textbf{90.8} & 90.6 &90.1\\
    SPHConv~\cite{lei2019octree} & 
        \textbf{90.3} & 4.1 & 90.0 & 90.0\\
    \bottomrule
\end{tabular}%
\end{minipage}
\hspace{0.025\textwidth}
\begin{minipage}[t]{0.475\textwidth}
\setlength{\tabcolsep}{4pt}
\centering
\caption{Comparison of accuracy obtained with our initialization against batch normalization for different batch sizes on the \dataset{ModelNet40} data set.}
\label{tbl:class_small_batch}
\begin{tabular}{crrrr}
    \toprule
    \multicolumn1r{Batch size $\rightarrow$} & \multicolumn{2}{c}{2}& \multicolumn{2}{c}{4}\\
    \cmidrule(lr){2-3}\cmidrule(lr){4-5}
    & Ours & BN & Ours & BN\\
    \cmidrule{1-5}
    PCCNN~\cite{atzmon2018pccnn} & 
        \textbf{90.3} & 28.3 & 90.3 & \textbf{90.4}\\ 
    KPConv~\cite{thomas2019KPConv} & 
        \textbf{90.3} & 14.6 & \textbf{90.6} & \textbf{90.6}\\ 
    MCConv~\cite{hermosilla2018mccnn}  & 
        \textbf{90.0} & 14.8 & 90.0 & \textbf{90.1}\\ 
    PointConv~\cite{wu2019pointconv}  & 
        \textbf{90.5} & 13.2 & \textbf{90.6} & 90.4\\ 
    SPHConv~\cite{lei2019octree}  & 
        \textbf{90.0} & 19.7 & \textbf{90.1} & 88.6\\ 
    \bottomrule
\end{tabular}%
\end{minipage}
\end{table}%

\subsection{Gradient accumulation}

We trained the two best performing convolutions on the \dataset{ScanNet} data set, \method{MCConv} and \method{KPConv}, with a virtual batch size of $4$. 
We sample each scene as described in paper’s Sect. 5.4, i.e., making each scene fill the GPU memory, and we accumulate the gradients over 4 scenes. 
In this setup, batch normalization still fails, obtaining an IoU of $32.1$ for \method{KPConv} and $54.5$ for \method{MCConv}.

\section{Implementation details}
\label{imp}

We implemented the different convolution operations in a unified framework using custom ops to reduce memory consumption.
Each method was implemented as a single custom op following the official repositories of the tested operations.
However, for some operations, we performed slight modifications.
In the following subsections, we will describe the individual changes.

\subsection{KPConv}

The proposed \method{KPConv} convolution operation by Thomas et al.~\cite{thomas2019KPConv} use a sum as aggregation method, $\estimator_\mathrm{Sum}$.
However, for the results reported in our paper, we substituted such method by a Monte Carlo estimation, $\estimator_\mathrm{MC}$.
This change was necessary since we experienced instability during training for small batch sizes.
We believe that the lack of normalization on the operator produces a high variance in the features between 3D models with dense neighborhoods, such as plants, and 3D models with neighborhoods with low density, such as objects from the class Table.
By using Monte Carlo estimation, $\estimator_\mathrm{MC}$, we were able to obtain a stable training for \method{KPConv} with low batch sizes.
Table~\ref{tbl:kpconv_abla} presents the accuracy of the original \method{KPConv} and our modified operation, \method{KPConv(N)}, for the \dataset{ModelNet40} data set with varying batch sizes.
In these results, we can see that our modification not only makes the training stable for small batches but also achieves higher accuracy than the original operation for large batches.
Moreover, we can see that our initialization scheme is also able to train the original \method{KPConv} operation for large batches.

\begin{table}
\centering
\caption{Comparison of accuracy obtained with our initialization against batch normalization for the two variants of \method{KPConv} on the \dataset{ModelNet40} data set.}
\vspace{.1cm}
\label{tbl:kpconv_abla}
\begin{tabular}{crrrrrr}
    \toprule
    \multicolumn1r{Batch size $\rightarrow$} & \multicolumn{2}{c}{2}& \multicolumn{2}{c}{4}& \multicolumn{2}{c}{8}\\
    \cmidrule(lr){2-3}\cmidrule(lr){4-5}\cmidrule(lr){6-7}
    & Ours & BN & Ours & BN & Ours & BN\\
    \cmidrule{1-7}
    KPConv & 
        4.1 & \textbf{21.4} & 4.1 & 90.4 & \textbf{90.9} & 91.0\\ 
    KPConv(N) & 
        \textbf{90.3} & 14.6 & \textbf{90.6} & \textbf{90.6} & 90.6 & \textbf{91.5}\\ 
    \bottomrule
\end{tabular}%
\end{table}%

\subsection{PointConv}

The \method{PointConv} convolution operator~\cite{wu2019pointconv} processes the relative position of the neighboring points with an MLP to produce the weights used to modulate the different features.
Unfortunately, when the receptive field of the convolutions is small the relative position is also close to zero, and when the receptive field of the convolution is too large the relative positions are also large.
This results in vanishing or exploding gradients for large-scale data sets such as \dataset{ScanNet}.
We follow the original implementation and use batch normalization between the layers of such MLP, which normalizes the input based on data statistics and enables training on the aforementioned data sets.
However, when batch normalization is not used another approach has to be considered.
Other convolutions such as \method{MCConv}~\cite{hermosilla2018mccnn} normalize the input by the radius of the receptive field.
However, in order to remain as close as possible to the original method, we used instead a custom initialization scheme for the first layer of this MLP:
\[
Var(w) = \frac{1}{d r^2}
\]
\noindent where $d$ is the number of dimensions and $r$ is the radius of the receptive field.
Please note that this initialization is only used in our setup when batch normalization is not considered.

\section{Other architectures}
The number of variables on neural network architectures for point clouds is large: convolution operation, pooling method, neighborhood selection, architecture, data augmentation, etc.
Each paper uses a different combination of those different variables which not only makes them difficult to compare, but could even mislead the reader on the improvements of novel contributions, such as our weight initialization scheme. 
Therefore, to strive for reproducibility, we favored a fixed architecture and only modified the convolution operation to see the improvement of our initialization on the different convolutions.
However, in this section, we further analyze the effects of our initialization scheme on a different architecture.
Moreover, we also compare our architecture with the different architectures proposed for each convolution operator.

\subsection{PointConv}
In order to measure the improvement of our initialization scheme on other architectures, we used it on the network proposed by \citet{wu2019pointconv}.
This architecture differs from ours in several parameters.
It uses farthest point sampling instead of Poisson disk sampling to reduce the size of the point cloud, it selects the neighboring points using the K-nearest neighbors algorithm, and uses several linear layers between each convolution.
We used the code from the official repository and trained a model in the \dataset{ScanNet} data set with batch normalization, a batch size equal to $8$, and a point cloud size of $6,000$.
Then, we trained the same model without batch normalization and our weight initialization algorithm, but used instead a point cloud size of $16,000$ and a batch size of $1$.
In order to virtually increase the batch size in our setup, we accumulated the gradients over $8$ training steps before modifying the weights of the model.
Note that both configurations filled up the memory of our graphics card.

This experiment showed similar results to the ones presented in the main paper.
The model with batch normalization obtained a IoU of $57.7$, whilst the model without it and our initialization achieved a IoU of $59.3$.
Moreover, when our weight initialization is not used, the same model is not able to train, achieving an IoU of $1.7$.
Lastly, the model with batch normalization trained with the larger point clouds and small batch size was able to achieve only an IoU of $30.5$.

\subsection{Architecture comparison}
Lastly, we compare the results obtained on our architecture with batch normalization and without our initialization, to the best results reported on the \dataset{ScanNet} validation set of the different methods. 
As shown in Tbl.~\ref{tbl:orig_arch} our architecture presents state-of-the-art results, obtaining better or similar results for most of the methods.

\begin{table}[h]
    \small 
    \setlength{\tabcolsep}{3pt}
    \centering
    \caption{ScanNet semantic segmentation comparison with our and original architectures, both with standard initialization and batch normalization.}
    \vspace{.1cm}
    \begin{tabular}{cccccc}
        \toprule 
         &
         \multicolumn1c{PCCNN~\cite{atzmon2018pccnn}$^{*}$}&
         \multicolumn1c{KPConv~\cite{thomas2019KPConv}}&
         \multicolumn1c{MCConv~\cite{hermosilla2018mccnn}}&
         \multicolumn1c{PointConv~\cite{wu2019pointconv}}&
         \multicolumn1c{SPHConv~\cite{lei2019octree}}\\
         \midrule
         Ours & \textbf{65.7} & 66.0 & \textbf{66.1} & \textbf{64.2} & 59.5\\
         Orig. & -- & \textbf{69.2} & 63.3 & 61.0 & \textbf{61.0}\\
         \bottomrule
    \end{tabular}
    {\\ \scriptsize $*$ This method was only tested on scenes composed of single models, e.g. ShapeNet.}
    \label{tbl:orig_arch}
\end{table}

\end{document}